\documentclass[conference]{IEEEtran}
\IEEEoverridecommandlockouts
\usepackage{cite}
\usepackage{amsmath,amssymb,amsfonts}
\usepackage{algorithmic}
\usepackage{graphicx}
\usepackage{textcomp}
\usepackage{xcolor}
\usepackage{comment}
\usepackage{float}
\usepackage{mathrsfs}  
\usepackage{hyperref}
\def\BibTeX{{\rm B\kern-.05em{\sc i\kern-.025em b}\kern-.08em
    T\kern-.1667em\lower.7ex\hbox{E}\kern-.125emX}}
\begin{document}

\title{A Learning-Based Framework for Collision-Free Motion Planning\\

}

\author{
\IEEEauthorblockN{Mateus Salomão, Tianyü Ren, Alexander König}
\IEEEauthorblockA{
Chair of Robotics and Systems Intelligence \\
Technical University of Munich, Munich, Germany \\
\{mateus.salomao, tianyu.ren, koenig\}@tum.de
}
}

\maketitle

\begin{abstract}
This paper presents a learning-based extension to a Circular Field (CF)-based motion planner for efficient, collision-free trajectory generation in cluttered environments. The proposed approach overcomes the limitations of hand-tuned force field parameters by employing a deep neural network trained to infer optimal planner gains from a single depth image of the scene. The pipeline incorporates a CUDA-accelerated perception module, a predictive agent-based planning strategy, and a dataset generated through Bayesian optimization in simulation. The resulting framework enables real-time planning without manual parameter tuning and is validated both in simulation and on a Franka Emika Panda robot. Experimental results demonstrate successful task completion and improved generalization compared to classical planners.
\end{abstract}

\
\section{Introduction}

With the increasing adoption of robot manipulators for collaborative tasks and industrial applications, the need for efficient and safe motion-planning algorithms has become ever more pressing.

Among classical approaches, the most common methods are sampling-based and field-based planners.

Sampling-based motion planners have been widely adopted due to their effectiveness and ease of implementation.
Nonetheless, these planners suffer from slow convergence and exhibit long run times in complex environments \cite{b1}.

In applications which require robots to operate in highly constrained or cluttered environments - such as inspection, material handling and assembly - field-based planning can be a more suitable alternative.

A pioneering method in this direction is the Artificial Potential Fields (APF) approach, which generates collision-free paths by combining repulsive obstacle forces with goal attraction \cite{b2}. Albeit computationally inexpensive, APF is prone to converging to local minima. 

A similar line of methods which addresses this issue is based on Circular Fields (CF) \cite{b6}. Such methods are local-minima free and the robot is guided around obstacles by the Lorentz forces resulting from artificial magnetic fields. 

Further variations have focused on defining the artificial current vector more consistently \cite{b7}, and under certain conditions, formal guarantees for collision avoidance and goal convergence have been established \cite{b8}. Although robust, these methods´ success is highly sensitive to the selection of a high number of non-intuitive gains and parameters related to the force fields.

In light of the aforementioned limitations, learning-based methods have been proposed. 

Reinforcement learning is popular for its ability to generalize to unknown environments: the agent learns a policy that maps states to optimal actions by directly interacting with the environment and receiving reward signals. 
With appropriately designed rewards, the motion-planning problem can be solved. However, the method´s sample complexity is a well-known limitation. \cite{b1}


A third avenue for solving the motion planning paradigm consists of using deep learning methods to enhance classical methods and improve their generalization and avoid manual parameter selection. 
Even though there are well-established approaches which increase the efficiency and robustness of sampling-based methods (\cite{b4}, \cite{b5}) there has not been much effort in the robotics research community to extend the idea to field-based planners.


The main contribution of this paper is a learning-based extension of the Predictive Multi-Agent CF-based trajectory planner from \cite{b9}, enabling it to generate feasible start-to-goal trajectories in real time from a single depth image of the scene without exhaustive parameter tuning. 

To generalize to arbitrary scenes, we represent the environment as an occupancy grid of spheres.
We collect a dataset of highly randomized scenes paired with optimal planner-parameter vectors in simulation and train a neural network to predict the appropriate parameters.



\section{Problem Formulation}

The motion-planning problem presented in this paper consists of finding a valid trajectory in the free task space \( \mathcal{T}^c \subset \mathbb{R}^3 \) from an initial position \( \vec{x}_0 \in \mathcal{T}^c \) to a goal position \( \vec{x}_g\), such that the trajectory avoids obstacles and satisfies joint and hard task constraints. 

Here, \( \mathcal{T}_{\text{obs}} \subset \mathbb{R}^3 \) denotes the region occupied by obstacles, and the free task space is given by \( \mathcal{T}^c = \mathcal{T} \setminus \mathcal{T}_{\text{obs}} \), where \( \mathcal{T} \subset \mathbb{R}^3 \) is the overall workspace.

\section{Circular Field Based Motion Planning}
In the planner formulation, reference trajectories for the robot end-effector are generated considering a point-mass dynamics model:
\begin{equation}
    m \ddot{\vec{x}}_d = \vec{F}_s
\end{equation}

where $\vec{x}_d$ is the desired acceleration of the robot control point, m denotes its virtual mass and $\vec{F}_s$ is the steering force:
\begin{equation}
\vec{F}_s = \vec{F}_{VLC} + k_{CF}\vec{F}_{CF} + k_{manip}\vec{F}_{manip} + k_r \vec{F}_r
\end{equation}

\subsection{Circular Field Force}
This component of the force field reproduces the effects of a virtual electromagnetic field on the robot, which is considered a charged particle in motion. 
Each obstacle $o_i$ is modeled as a sphere and assigned a current vector $\vec{c}_i$ which passes through its center. It then generates an artificial magnetic field component on the agent defined as $\vec{B}_i = k_{CF} \vec{c}_i \ \times \dot{\vec{x}}_i$
The CF force produced on the agent by the i-th obstacle is then based on Lorentz´s law (as illustrated in Fig. 1):
\begin{equation}
    \vec{F}_{CF, i} = \dot{\vec{x}}_i \ \times \vec{B}_i
\end{equation}
The authors of \cite{b9} define a detection shell with radius $r_d$ around each obstacle, such that $\vec{F}_{CF, i} = 0$ if $||\vec{x}_{o_i} - \vec{x}|| > r_d$.  
The resulting circular field force from $n_o$ obstacles  is given by
\begin{equation}
    \vec{F}_{CF} = \sum_{j=0}^{n_o}\vec{F}_{CF, i}
\end{equation}

\begin{figure}[H]
    \centering
    \includegraphics[width=\linewidth]{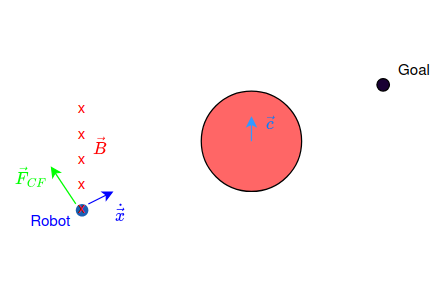}
    \caption{Scheme illustrating the circular field (dark red) and CF force (green) generation for a
spherical obstacle (light red) in 2D.}
    \label{fig:circular_field_scheme}
\end{figure}

\subsection{Attractive force}
This component is responsible for attracting the robot towards the goal pose:
\begin{equation}
    \vec{F}_{VLC} = -k_v \times (\dot{\vec{x}} - \nu \dot{\vec{x}}_d)
\end{equation}
with $\dot{\vec{x}}_d = \frac{k_p}{k_v}(\vec{x}_g - \vec{x})$

\subsection{Manipulability force}
A third force field component is introduced to improve the planner´s robustness with respect to manipulability, undesired contacts and joint limits (\cite{b9}):
\begin{equation}
    \vec{F}_{manip} = \lambda \vec{w}_{min}
\end{equation}
where $\vec{w}_{min}$ is the singular vector associated with the singular minimum value $\sigma_{min}$ considering the singular value decomposition of the task space velocity Jacobian $\vec{J}_{\vec{p}_\mathcal{X}}$ 

\subsection{Repulsive Potential Field Force}
Following \cite{b2,b9}, we include APF repulsive forces:
\begin{equation}
    \vec{F}_{r, i} = k_r \, \vec{d}_i \left( \frac{1}{\|\vec{d}_i\|} - \frac{1}{r_d} \right) \frac{1}{\|\vec{d}_i\|^2}
\end{equation}

where $k_r$ is the repulsive force gain and $\vec{d}_i = \vec{x}_{o_i} - \vec{x}$ denotes the vector from the robot control point to the closest obstacle surface point. Similarly to the CF component, this force is null if $d_i < r_d$.

The total repulsive force is then:
\begin{equation}
    \vec{F}_r = \sum_{i = 1}^{n_o} \vec{F}_{r, i}
\end{equation}

\subsection{Obstacle representation and perception pipeline}
In contrast to previous experiments, where scenarios were simplified by modeling obstacles as spheres with predefined radii and centers, our pipeline must handle an arbitrary number of obstacles with varied shapes. To achieve this, the perception module converts a raw depth image into an occupancy grid of spherical voxels, effectively capturing the 3D structure of the scene for downstream planning.
\section{Multi Agent Framework}

During a planning step, \(n_a\) predictive agents are generated to explore the environment (here \(n_a = 7\)). For each agent \(i\), the obstacles \(o_j^i\), \(j=1,\dots,N\), are assigned current vectors \(\vec{c}_i\). These vectors are generated by different heuristics that influence how each obstacle is avoided.

For example, the velocity heuristic yields an avoidance direction which is close to the current direction of the robot motion.

The path length heuristic, in turn,  aims to
minimize the path length for circumventing an obstacle.

Further heuristics are constructed by combining the effects of the previous ones, which leads to for example the goal-vector, obstacle-distance, path length-obstacle distance and random artificial current \cite{b9}.

Each agent is evaluated at equidistant sampling times,
the agent with the lowest cost is selected and its parameters
are copied by the real robot. 

The cost function for each agent is
\[
\tilde{\mathcal{F}} = c_{\text{pl}} + c_{\text{gd}} + c_{\text{od}} + c_{\text{ws}},
\]
with terms:
\begin{itemize}
  \item Path length:
  \[
    c_{\text{pl}} = w_{\text{pl}} \sum_{i=t+1}^{t+T_p} \|\vec{x}(i) - \vec{x}(i-1)\|.
  \]
  \item Distance to goal:
  \[
    c_{\text{gd}} = w_{\text{gd}} \,\|\vec{x}_g - \vec{x}(t+T_p)\|.
  \]
  \item Obstacle proximity:
  \[
    d_{\min} = \min_{\substack{i \in \{t+1,\dots,t+T_p\} \\ j \in \{1,\dots,n_o\}}} \|\vec{x}_{o_j}(i) - \vec{x}(i)\|,\quad
    c_{\text{od}} = \frac{w_{\text{od}}}{d_{\min}}.
  \]
  \item Workspace-boundary violation:
  \[
    c_{\text{ws}} = w_{\text{ws}} \sum_{i=t+1}^{t+T_p} \sum_{k=1}^{3}
    \begin{cases}
      (\vec{x}_k(i) - p_k^+)^2 & \text{if } \vec{x}_k(i) > p_k^+,\\
      (\vec{x}_k(i) - p_k^-)^2 & \text{if } \vec{x}_k(i) < p_k^-,\\
      0 & \text{otherwise.}
    \end{cases}
  \]
\end{itemize}

While the predictive agents are
being simulated, the real robot is moving under the influence
of the CF and the VLC using the parameters of the current
best agent instead of just following the trajectory of this agent

\section{Learning-Based Pipeline}

As previously mentioned, the performance of the described motion planner is strongly dependent on the selection of a number of hyperparameters, in particular the position, velocity and electromagnetic force gains.
If not tuned properly, the planner might fail to generate a collision-free trajectory towards the goal, as will be seen in  the results.

To overcome this limitation and eliminate the need for manual parameter selection in every scene, we extend the approach with a deep learning–based parameter inference.
Hence, we define the planner parameter vector $\vec{p} = [\vec{k}_p^T, \vec{k}_v^T, \vec{k}_{CF}^T, \vec{k}_{manip}^T, \vec{k}_{r}^T, r_d] $ as our model output given an arbitrary input scene.

Note that we do not impose equal gains for all agents, recognizing that the different heuristics may require different parameters.

A schematic overview of the proposed pipeline is presented below.

\begin{figure}[H]
    \centering
    \includegraphics[width=\linewidth]{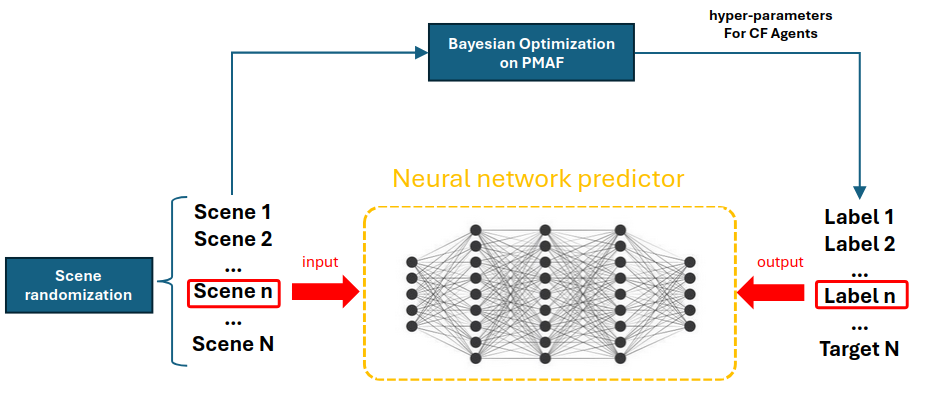}
    \caption{System diagram of the proposed learning-based approach for collision-free motion planning. Randomized scenes are used to generate optimal hyperparameter labels via Bayesian optimization on robot PMAF-generated trajectory costs. These labeled pairs are used to train a neural network that predicts suitable hyperparameters for novel input scenes.}
    \label{fig:circular_field_scheme}
\end{figure}

\subsection{Model inputs}
The inputs \(\{\vec{X}_n\}_{n=1}^N\) are the point clouds of all \(N\) scenes - subsampled to 2,500 points and expressed in the robot’s base frame - directly converted from the depth images.

\subsection{Model outputs}
The output set $ \{ \vec{y}_n\}_{1, ..., N} $ consists of the parameter vectors which lead to the optimal planner performance in scene n: $\vec{y}_n = \vec{p^*_n}$.

One key premise for a successful supervised deep learning method is the quality of the dataset. This aspect posed a significant challenge in the current framework, as it is not possible to obtain ground truth values  $\vec{p^*}$ for an arbitrary scene n.

The dataset labeling can thus be framed as a hyperparameter optimization problem of the form:
\begin{equation}
    \vec{p^*} = \operatorname*{arg\,min}_{\vec{p}} f(\vec{p})
\end{equation}

in which:
\begin{itemize}
    \item $\vec{p}$ are the hyperparameters with domain or search space $\mathbb{R}^{36}$
    \item $f(\vec{p}) = J_{PMAF}(\vec{p})$ is the cost of the robot trajectory from start to goal obtained by running the PMAF planner parametrized with $\vec{p}$ 
\end{itemize}

\subsubsection{Cost Function Design}

Firstly, it was necessary to design $J_{PMAF}$ to meaningfully assess the success of the robot in completing the task given its Cartesian path $\mathcal{P} = (\vec{x}_0, \vec{x}_1, ..., \vec{x}_T)$, where $\vec{x}_t$ is the robot end-effector´s position at timestep t and T is the final timestep. We define:
\[
    J_{\mathrm{PMAF}} = w_{\text{cl}}\,C_{\text{cl}} + w_{\text{pl}}\,C_{\text{pl}} + w_{\text{sm}}\,C_{\text{sm}} + w_{\text{gd}}\,C_{\text{gd}},
\]
with
\begin{itemize}
    \item \(C_{\text{cl}} = \frac{1}{T}\sum_{t=1}^{T}\frac{1}{d_{\min,t}}\), where \(d_{\min,t}\) is the minimum distance to any obstacle at \(t\),
    \item \(C_{\text{pl}} = \sum_{t=0}^{T-1}\|\vec{x}_{t+1}-\vec{x}_t\|\) (total path length),
    \item \(C_{\text{sm}} = \frac{1}{T-1}\sum_{t=2}^{T-1}\|\vec{x}_{t+1}-2\vec{x}_t+\vec{x}_{t-1}\|^2\) (smoothness cost),
    \item \(C_{\text{gd}} = \|\vec{x}_T - \vec{x}_g\|\) (goal deviation).
\end{itemize}

\subsubsection{Optimization technique}

We solve the hyperparameter tuning problem 
\[
    \vec{p^*} \;=\;\arg\min_{\vec p\in\mathbb R^{36}} f(\vec p),
    \quad f(\vec p)=J_{\mathrm{PMAF}}(\vec p)
\]
using Bayesian optimization (BO), which comprises three main steps:

\begin{enumerate}
  \item Surrogate fitting: fit a Gaussian process \(p(y\mid\vec{p},D_{t-1})\) to samples \(D_{t-1} = \{(\vec{p}_i,f_i)\}\).  
  \item Acquisition maximization: choose \(\vec{p}_t\) that maximizes the acquisition function \(a(\vec{p})\).  
  \item Data augmentation: evaluate \(f_t = f(\vec{p}_t)+\mathcal{N}(0,\sigma^2)\) and append \((\vec{p}_t,f_t)\) to \(D_{t-1}\).
\end{enumerate}

The acquisition function plays an important role in BO, as it trades off exploration (searching for configuration space regions where the GP model has high associated uncertainty) and exploitation (which attempts to find $\vec{p}$ near the global minimum).

Repeating these steps yields rapid convergence to the global minimum under smoothness and curvature conditions, outperforming random or grid search in sample efficiency.

Despite this, classical BO comes with a few limitations, in particular \cite{b10}:
\begin{itemize}
  \item Homoscedasticity \& stationarity assumptions:  Real‐world hyperparameter landscapes often exhibit input‐dependent noise and non‐stationary behavior, causing standard GPs to misestimate uncertainty.
  \item Single‐objective acquisitions: A single criterion (EI, LCB, etc.) can conflict with others, leading to suboptimal sampling decisions.
\end{itemize}

To address these issues, we employ HEBO \cite{b10}, which enhances classical BO by introducing a flexible surrogate model that handles heteroscedasticity and non-stationarity via input/output transformations. It also leverages multi-objective acquisition through Pareto optimization of a portfolio of acquisition functions, and ensures robust candidate selection using trust regions and adaptive noise calibration.

Empirical studies demonstrate that these combined improvements yield faster convergence and higher median performance across diverse black‐box benchmarks.

\subsubsection{Neural network architecture}
The neural network used for gain prediction is based on PointNet++ \cite{b11}. 

The main reason behind the selection of this architecture is the model´s capability to summarize an input point cloud by a sparse set of
key points, roughly corresponding to the skeleton of
objects. We thus expected the model to be able to transmit this understanding to its outputs (i.e. more tightly constrained scenes probably require lower gains the

The core idea behind the vanilla PointNet network in \cite{b12} is to learn a spatial encoding of each point and then aggregate all individual point features to a global point cloud signature.

Given an unordered point set $\{\vec{x}_1, \vec{x}_2, ..., \vec{x}_n\}$ with $\vec{x}_i \in \mathbb{R}^d$, one can define a set function $f : \vec{X} \rightarrow \mathbb{R}$ that maps a set of points to a vector:
\begin{equation}
    f(\vec{x}_1, \vec{x}_2, ..., \vec{x}_n) = \gamma \left( \underset{i=1,...,n}{\mathrm{MAX}} \; \{ h(\vec{x}_i) \} \right)
\end{equation}

where $\gamma$ and $h$ are usually multi-layer perceptron (MLP) networks.

\begin{figure}[H]
    \centering
    \includegraphics[width=\linewidth, height = 6cm]{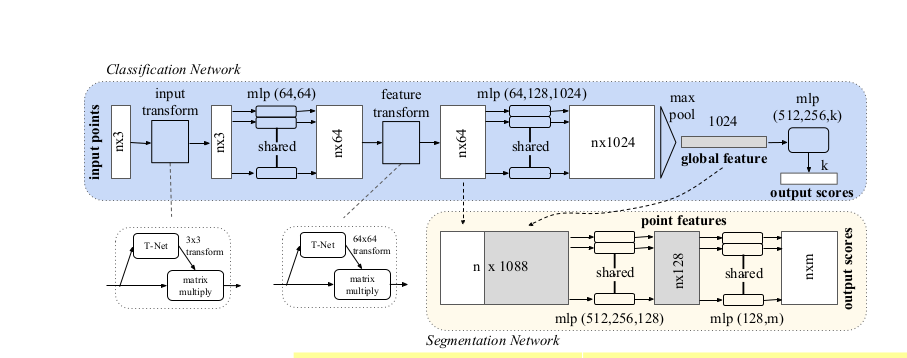}
    \caption{PointNet Architecture. The classification network takes n points as input, applies input and feature transformations, and then
aggregates point features by max pooling. The output is classification scores for k classes}
    \label{fig:circular_field_scheme}
\end{figure}

Here we modified the classification layer subsequent to the last MLP layer by a fully connected layer as we wish to regress the 36 dimensional vector $\vec{p^*}$

In order to guarantee invariance to rigid transformations, a separate mini-network (T-net in figure 3) predicts and applies an affine transformation matrix to the coordinates of input points \cite{b12}.

The authors of PointNet++ argue that exploiting local structure along the hierarchy improves model generalization and augments the basic architecture in figure 3 by adding sampling and grouping layers.   
More specifically, the updated architecture stacks \textit{set abstraction
modules}, composed of a Sampling layer, a Grouping layer and a PointNet layer. The Sampling
layer selects a set of points from input points, which defines the centroids of local regions. Grouping
layer then constructs local region sets by finding “neighboring” points around the centroids. PointNet
layer uses a mini-PointNet to encode local region patterns into feature vectors.
This updated architecture is shown below.

\begin{figure}[H]
    \centering
    \includegraphics[width=\linewidth]{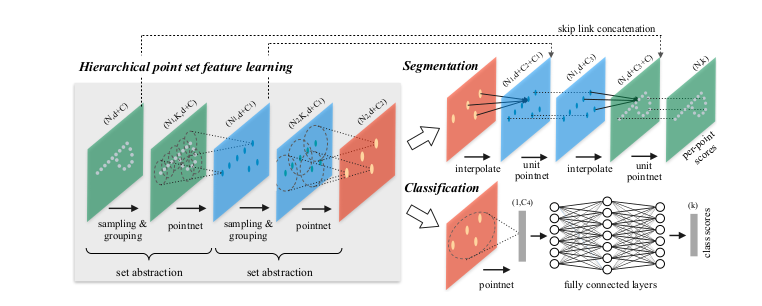}
    \caption{PointNet ++ Architecture. 
    The adapted model does not use a segmentation layer and replaces the final classification layer with k output classes by a fully connected layer with 36 output neurons}
    \label{fig:circular_field_scheme}
\end{figure}

\section{Results}

\subsection{Simulation}
The dataset labeling described in Section VI-B was implemented in the Genesis simulation engine. Fifty-six iterations of the optimization in Eq. 9 were applied to 1,000 randomized scenes. As the optimization was not successful for every scene (e.g., when obstacles were placed too close to the goal or starting poses), the task was considered successful and the corresponding pair $\{ \vec{X}_n, \vec{p^*_n}\}$ was only appended to the dataset if the robot reached the goal position within a neighborhood of $\epsilon = 0.03$ m. This criterion led to the selection of a subset of successful trajectories for the final dataset.

The scene randomization process consisted of adding fixed objects (a pillar next to the robot, a wall, and a desk) as well as a variable number of floating obstacles surrounding the robot. These obstacles consisted of 5 to 10 primitive shapes - cuboids, spheres, and cylinders. The positions of the objects, as well as the radii of the cylinders and spheres, were randomized across different scenes. Care was taken to avoid any intersections between the robot and the obstacles. The goal position also varied across scenes.

For simulation, a desktop running Ubuntu 24 and equipped with an NVIDIA RTX 3090 GPU was used.

Simulation results for two different scenarios are shown in Figures 5 and 6. The first and last frames of each trajectory are illustrated.
\begin{figure}[H]
    \centering
    \begin{minipage}[b]{0.49\linewidth}
        \centering
        \includegraphics[width=\linewidth]{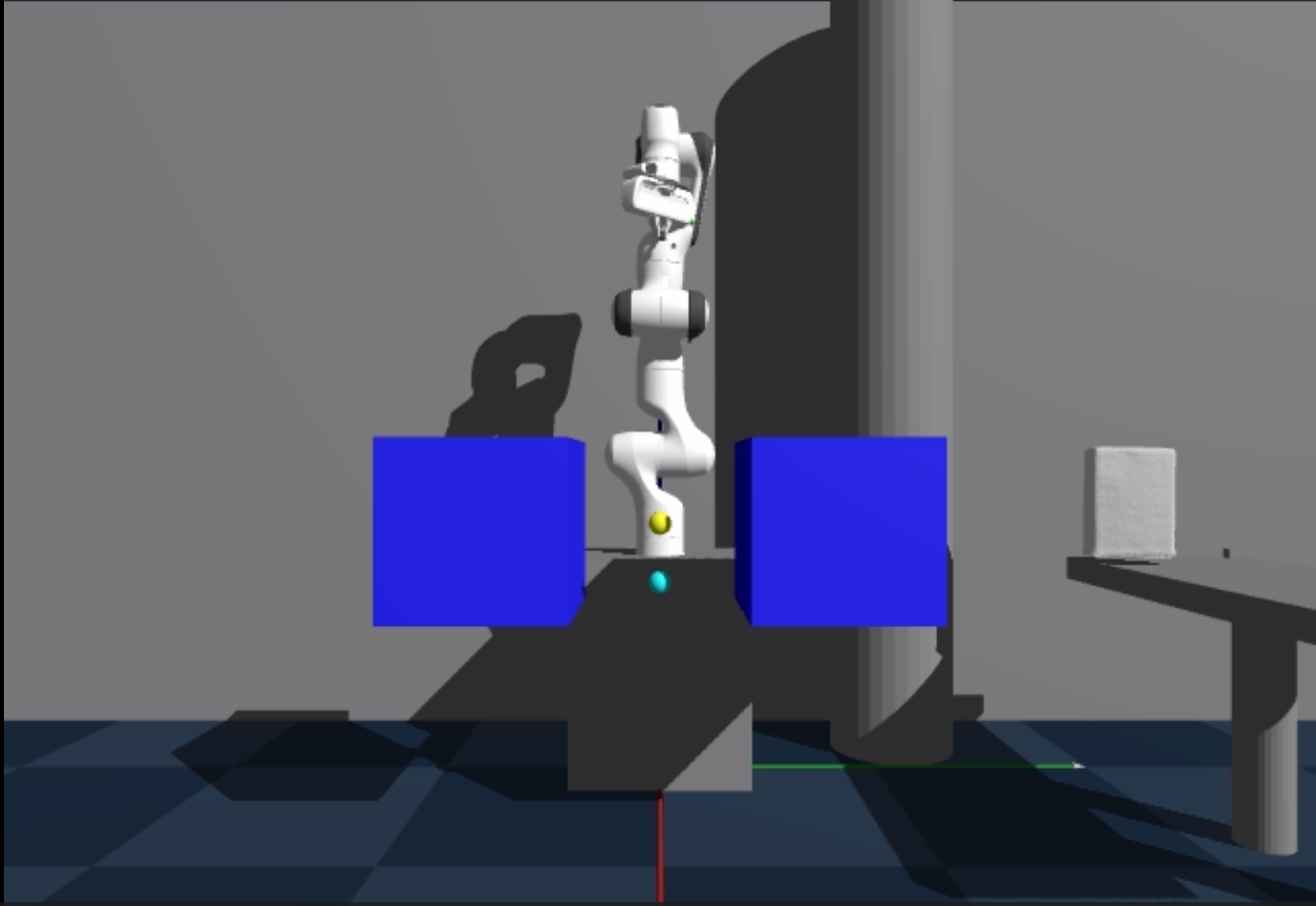}
    \end{minipage}
    \hfill
    \begin{minipage}[b]{0.49\linewidth}
        \centering
        \includegraphics[width=\linewidth]{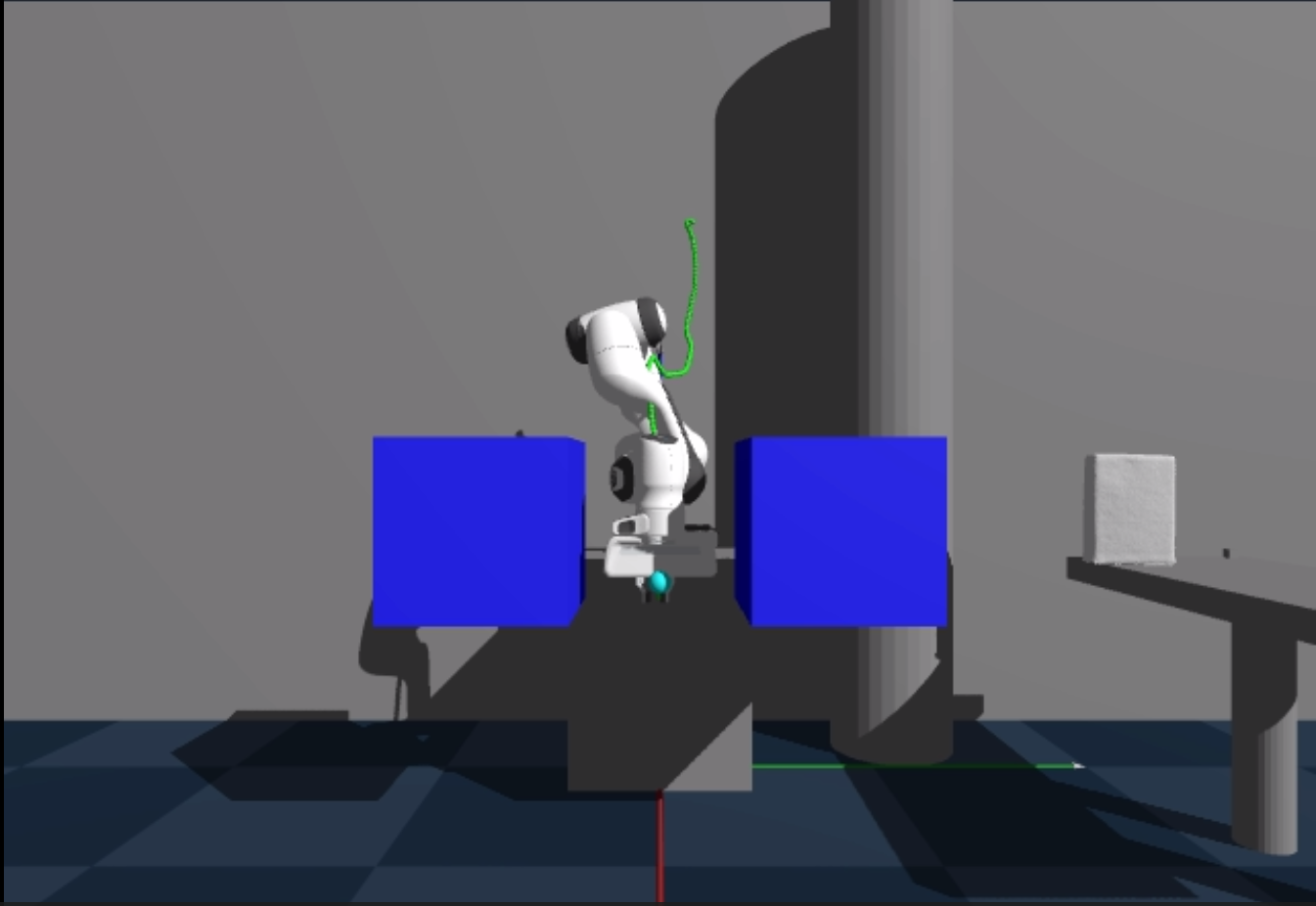}
    \end{minipage}
    \caption{First and last frames of the simulation. These illustrate the initial and final configurations of the task. The goal position for the robot’s flange and TCP are shown in yellow and light blue, respectively. TCP trajectories over time are shown in green.}
    \label{fig:video1}
\end{figure}

\vspace{-0.5em}

\begin{figure}[H]
    \centering
    \begin{minipage}[b]{0.49\linewidth}
        \centering
        \includegraphics[width=\linewidth]{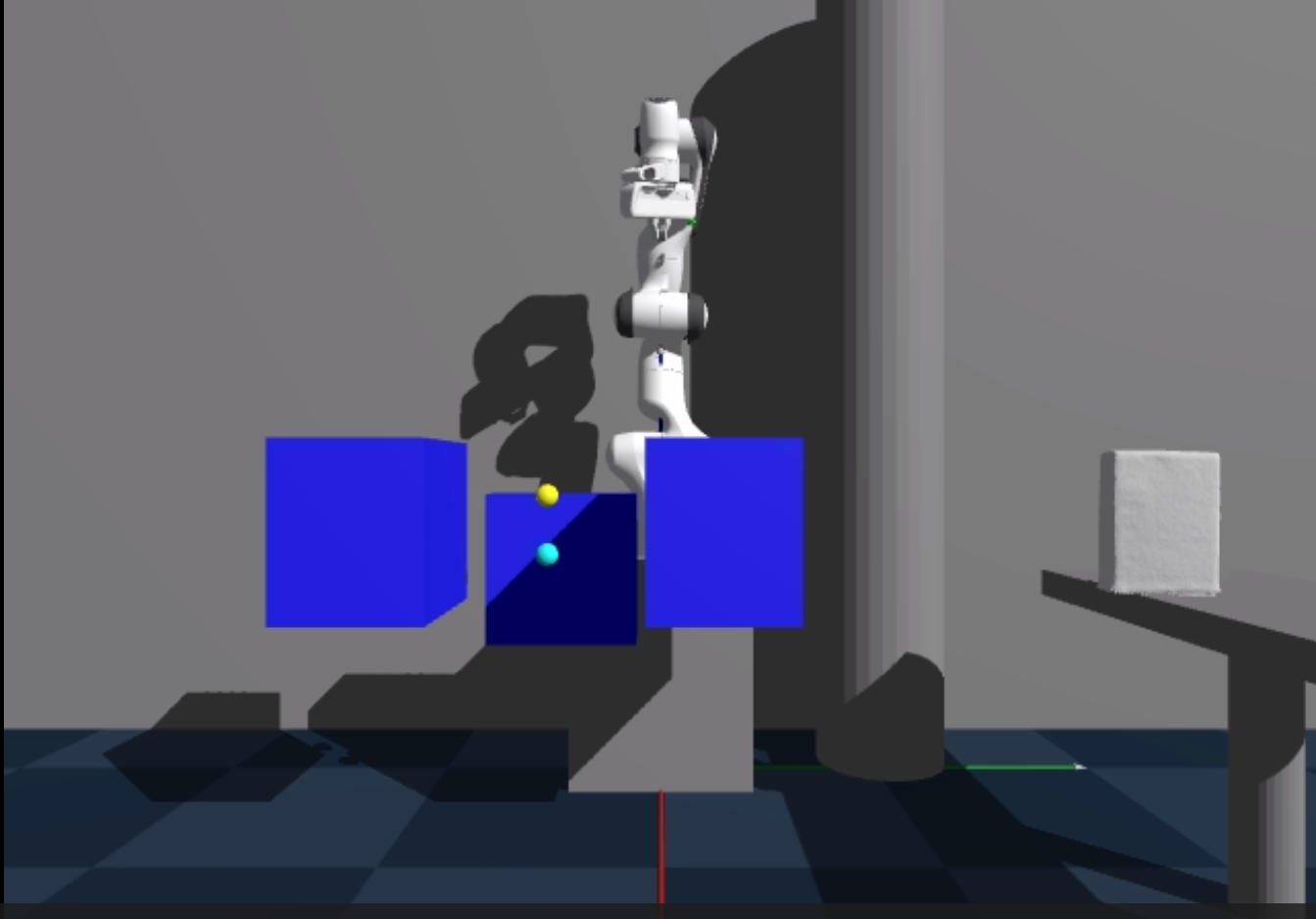}
    \end{minipage}
    \hfill
    \begin{minipage}[b]{0.49\linewidth}
        \centering
        \includegraphics[width=\linewidth]{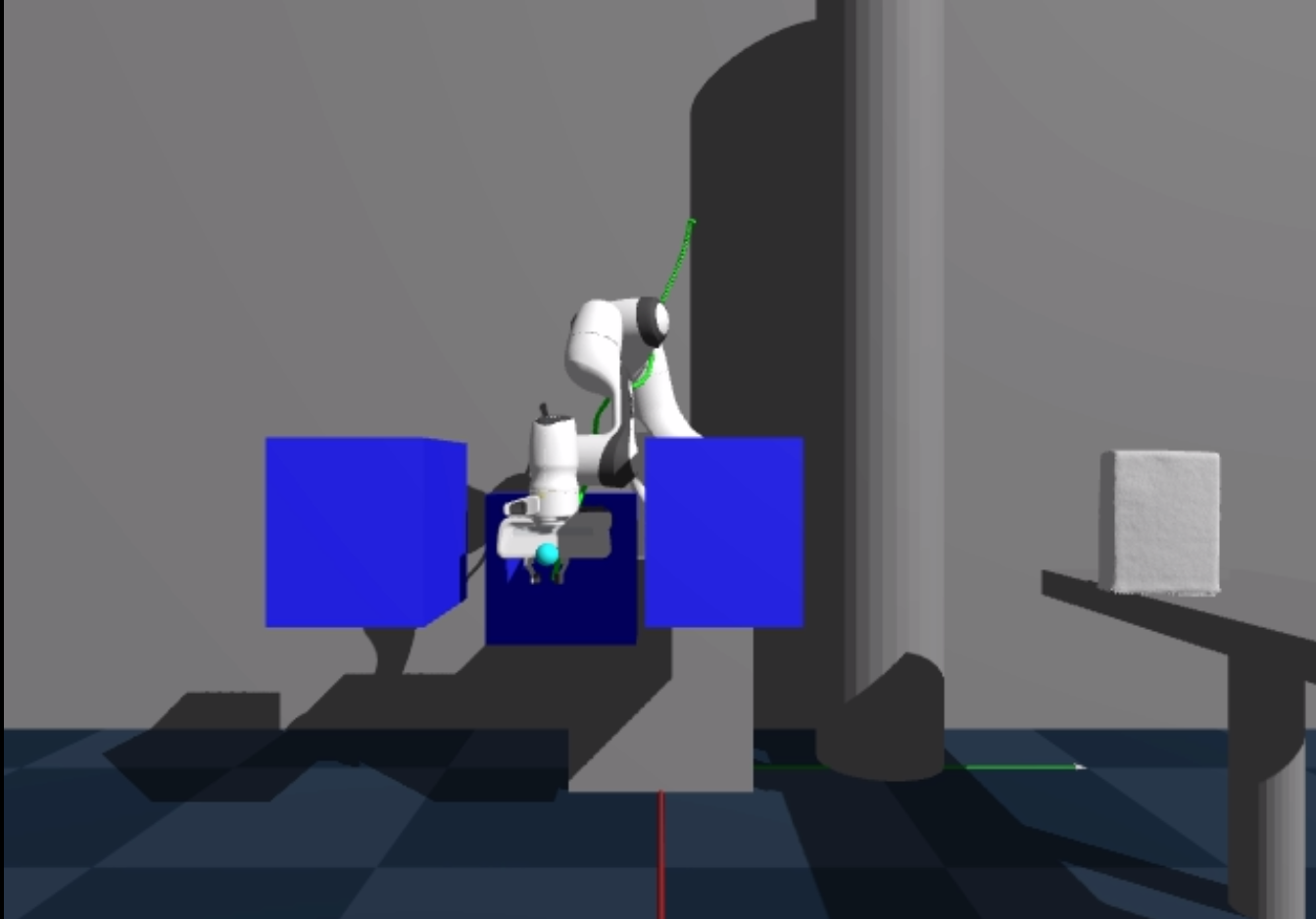}
    \end{minipage}
    \caption{First and last frames of the second simulation. Setup and visual encoding as in Figure~\ref{fig:video1}.}
    \label{fig:video2}
\end{figure}


\subsection{Robot experiments}
An Intel D-435 Realsense camera was used to generate the depth stream required by the perception pipeline, which processes only the first frame. Cartesian impedance control was implemented using the `panda-py` library \cite{b14}. The control loop runs at 1 kHz, taking Cartesian goal points from the PMAF planner and converting them into joint torques, which are transmitted to the Franka robot.

The planner node is fully implemented in C++ and runs at a base frequency of 10 Hz. The perception, control, and planning nodes exchange data via the Robot Operating System (ROS). The desktop PC is connected to the internal Franka controller via Ethernet.

Figures 7 and 8 show the results of horizontal and vertical obstacle avoidance tasks, respectively

\begin{figure}[H]
    \centering
    \begin{minipage}[b]{0.32\linewidth}
        \centering
        \includegraphics[width=\linewidth]{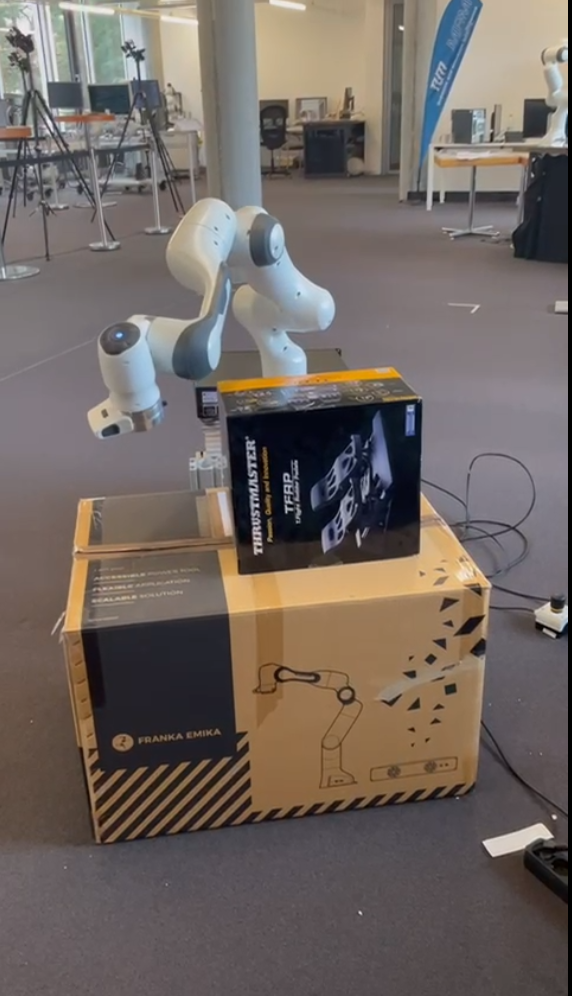}
    \end{minipage}
    \hfill
    \begin{minipage}[b]{0.32\linewidth}
        \centering
        \includegraphics[width=\linewidth]{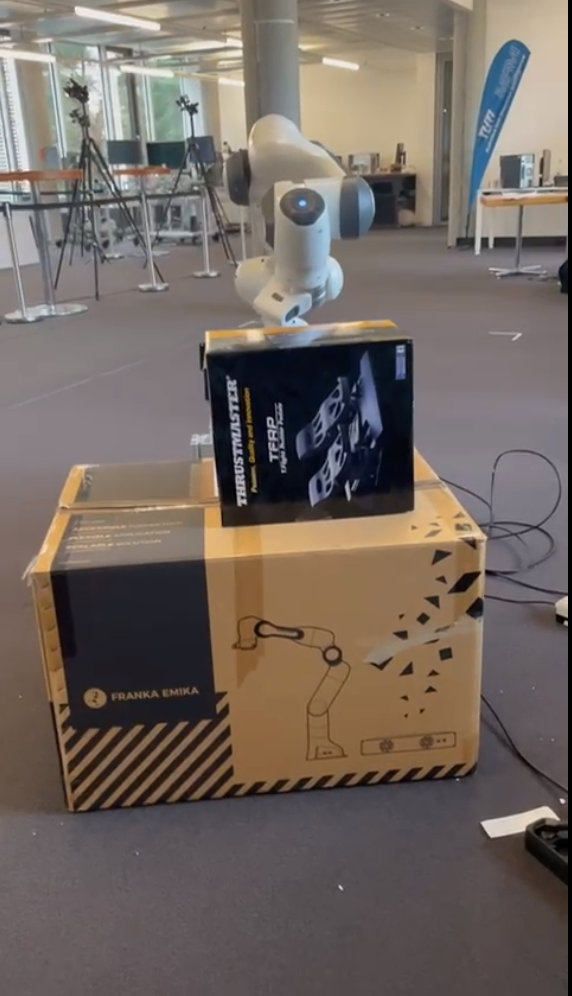}
    \end{minipage}
    \hfill
    \begin{minipage}[b]{0.32\linewidth}
        \centering
        \includegraphics[width=\linewidth]{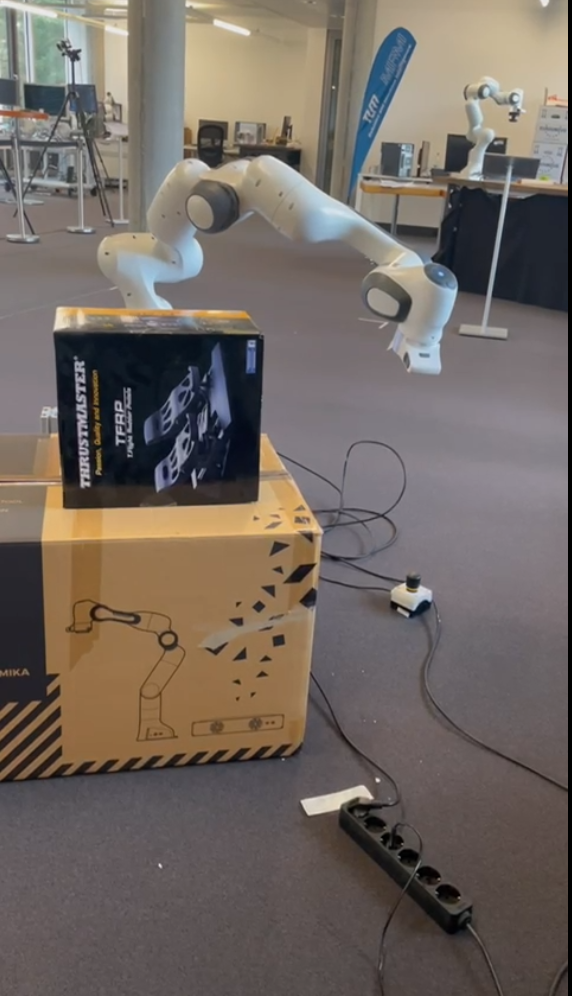}
    \end{minipage}
    \caption{First, intermediate, and last frames of the horizontal obstacle avoidance experiment. This sequence illustrates the robot's motion progression from initialization to task completion.}
    \label{fig:video_triplet}
\end{figure}

\begin{figure}[H]
    \centering
    \begin{minipage}[b]{0.32\linewidth}
        \centering
        \includegraphics[width=\linewidth]{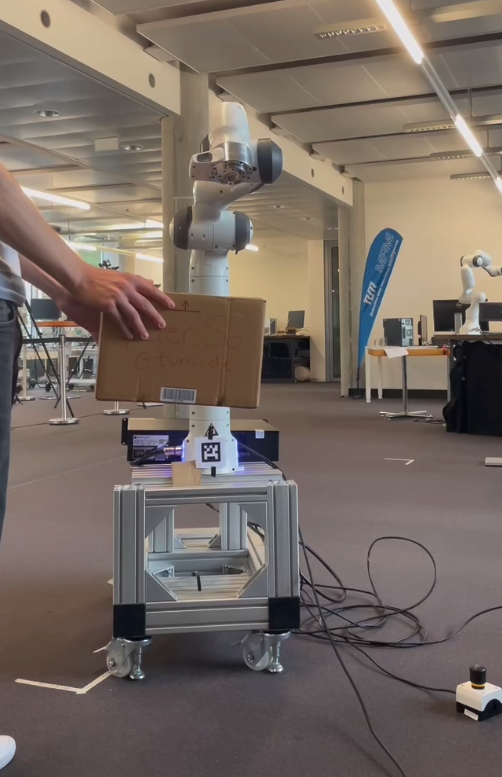}
    \end{minipage}
    \hfill
    \begin{minipage}[b]{0.32\linewidth}
        \centering
        \includegraphics[width=\linewidth]{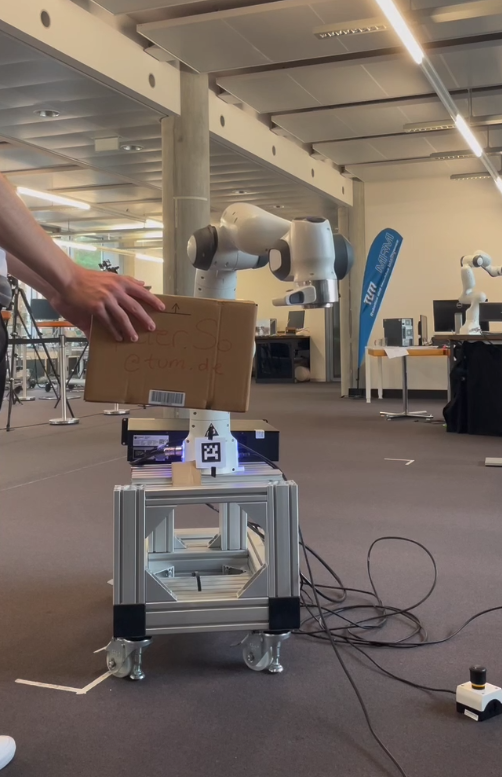}
    \end{minipage}
    \hfill
    \begin{minipage}[b]{0.32\linewidth}
        \centering
        \includegraphics[width=\linewidth]{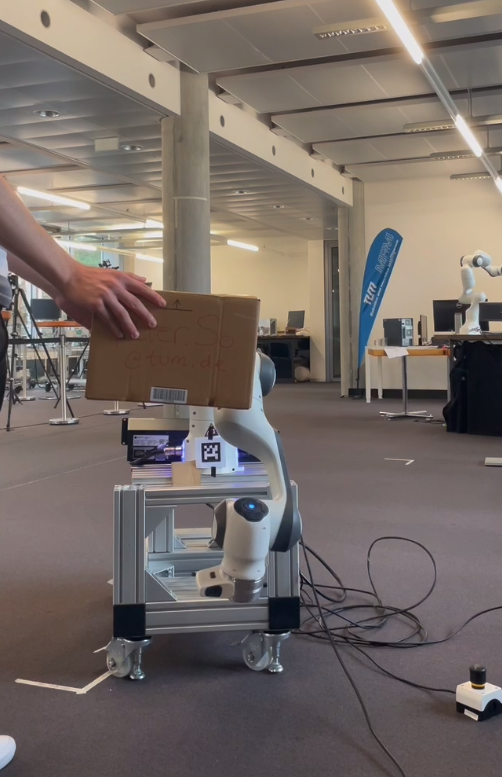}
    \end{minipage}
    \caption{First, intermediate, and last frames of the vertical obstacle avoidance experiment. This sequence illustrates the robot's motion progression from initialization to task completion.}
    \label{fig:video_triplet}
\end{figure}

The trajectory of the robot´s TCP, as well as the start and goal points, are shown below.

\begin{figure}[H]
    \centering
    \begin{minipage}[b]{0.49\linewidth}
        \centering
        \includegraphics[width=\linewidth]{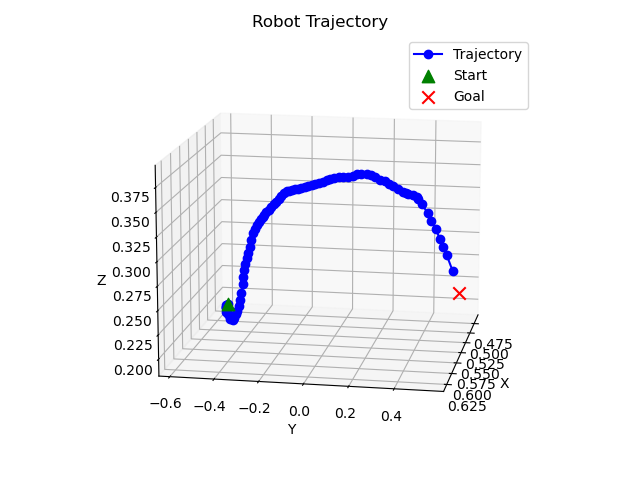}
    \end{minipage}
    \hfill
    \begin{minipage}[b]{0.49\linewidth}
        \centering
        \includegraphics[width=\linewidth]{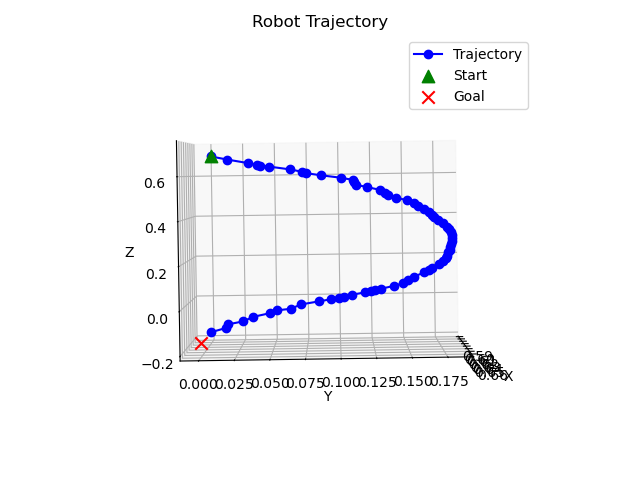}
    \end{minipage}
    \caption{Robot TCP trajectories for the horizontal (left) and vertical (right) obstacle avoidance experiments.}
    \label{fig:video2}
\end{figure}

In order to clarify the necessity of the gain prediction framework introduced in the present work, the gain inference was disabled and the original PMAF planner was run with standard pre-defined gains by the authors of \cite{b9}. The robot is unable to successfully reach the target positions in this case, colliding with the boxes. The corresponding results are shown in the following:

\begin{figure}[H]
    \centering
    \begin{minipage}[b]{0.49\linewidth}
        \centering
        \includegraphics[width=\linewidth]{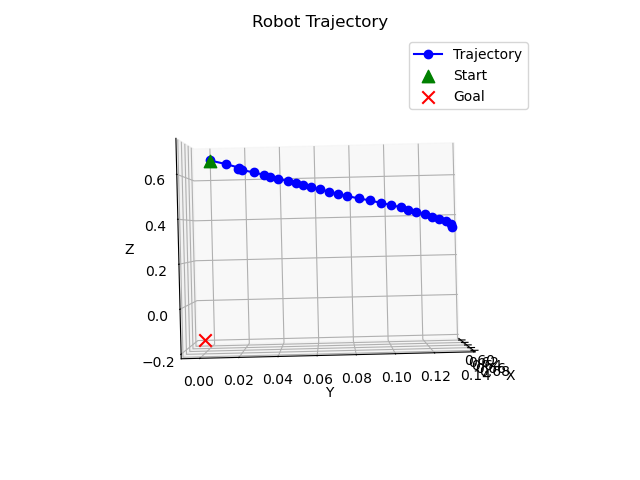}
    \end{minipage}
    \hfill
    \begin{minipage}[b]{0.49\linewidth}
        \centering
        \includegraphics[width=\linewidth]{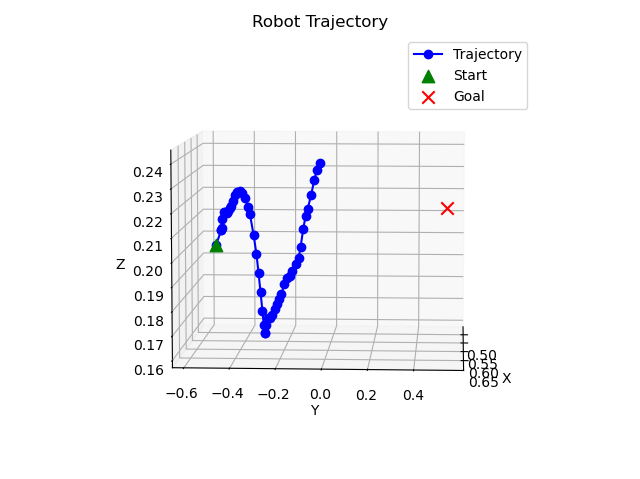}
    \end{minipage}
    \caption{Robot TCP trajectories for the horizontal (left) and vertical (right) obstacle avoidance experiments.}
    \label{fig:video2}
\end{figure}

For the corresponding animations we refer the reader to the project´s \href{https://github.com/mateusbsal4/neural-arm-field-planner}{GitHub repository}.

\section{Discussion and Conclusion}

In this paper, we propose a learning-based approach that infers the optimal parameter vector for a circular-field-based motion planner. Furthermore, we demonstrate the insufficiency of the original planner when predefined gains are used.

We also highlight the efficiency of the proposed method in terms of planning time: the total time taken by the robot to reach the goal was 12 s and 16 s for the horizontal and vertical obstacle avoidance tasks, respectively. In contrast, sampling-based planners often incur high computational costs - often reaching or exceeding 60 s - especially in high-dimensional environments \cite{b1}, \cite{b3}.

It is also worth noting that classical planners such as RRT-Connect were unable to generate valid paths for some of the simulation scenarios shown above.


Regarding future work, the first improvement would be to make the end-to-end planning pipeline reactive to dynamic obstacles. The current framework processes only the first frame of the depth stream and thus only supports static environments.

Another limitation concerns the resolution of the spherical voxel grid. Currently, each voxel has a 3 cm radius. Increasing the resolution would impose a heavy computational load on the desktop CPU and could violate the real-time constraints of the controller. Porting the current CPU-based planner implementation to CUDA - especially the voxel grid reading process - would greatly improve performance, as the planner currently stores the grid contiguously in CPU memory.

\vspace{12pt}


\begin{thebibliography}{00}
\bibitem{b1} M. G. Tamizi, M. Yaghoubi, and H. Najjaran, ``A review of recent trend in motion planning of industrial robots,'' \textit{Int. J. Intell. Robot. Appl.}, vol. 7, pp. 253--274, 2023. DOI: \href{https://doi.org/10.1007/s41315-023-00274-2}{10.1007/s41315-023-00274-2}.

\bibitem{b2} O. Khatib, ``Real-time obstacle avoidance for manipulators and mobile robots,'' \textit{Int. J. Robot. Res.}, vol. 5, no. 1, pp. 90--98, 1986.

\bibitem{b3} M. Becker, T. Lilge, M. A. Müller, and S. Haddadin, ``Circular fields and predictive multi-agents for online global trajectory planning,'' \textit{IEEE Robot. Autom. Lett.}, vol. 6, pp. 2618--2625, 2021. DOI: \href{https://doi.org/10.1109/LRA.2021.306199}{10.1109/LRA.2021.306199}.

\bibitem{b4} A. H. Ureshi and M. C. Yip, ``Deeply informed neural sampling for robot motion planning,'' in \textit{Proc. IEEE/RSJ Int. Conf. Intell. Robots Syst. (IROS)}, pp. 6582--6588, 2018.

\bibitem{b5} K.-C. Ying, P. Pourhejazy, C.-Y. Cheng, and Z.-Y. Cai, ``Deep learning-based optimization for motion planning of dual-arm assembly robots,'' \textit{Comput. Ind. Eng.}, vol. 160, p. 107603, 2021.

\bibitem{b6} L. Singh, H. Stephanou, and J. Wen, ``Real-time robot motion control with circulatory fields,'' in \textit{Proc. IEEE Int. Conf. Robot. Autom.}, vol. 3, pp. 2737--2742, 1996.

\bibitem{b7} S. Haddadin, R. Belder, and A. Albu-Schäffer, ``Dynamic motion planning for robots in partially unknown environments,'' \textit{IFAC Proc. Vol.}, vol. 44, no. 1, pp. 6842--6850, 2011.

\bibitem{b8} M. Becker, J. Köhler, S. Haddadin, and M. A. Müller, ``Motion planning using reactive circular fields: A 2D analysis of collision avoidance and goal convergence,'' \textit{IEEE Trans. Autom. Control}, early access, 2023. [Online]. Available: \href{https://doi.org/10.1109/TAC.2023.3303168}{https://doi.org/10.1109/TAC.2023.3303168}.

\bibitem{b9} R. Laha, M. Becker, J. Vorndamme, J. Vrabel, L. F. C. Figueredo, and M. A. Müller, ``Predictive multi-agent-based planning and landing controller for reactive dual-arm manipulation,'' \textit{IEEE Trans. Robot.}, vol. 40, 2024. [Online]. Available: \url{https://doi.org/10.1109/TRO.2024.3386638}

\bibitem{b10} 
R. Tutunov, A. Grosnit, R.-R. Griffiths, J. Wang, \textit{et al.}, ``An empirical study of assumptions in Bayesian optimisation,'' \textit{Preprint}, Feb. 2021. [Online]. Available: \url{https://www.researchgate.net/publication/349305527}

\bibitem{b11} 
C. R. Qi, L. Yi, H. Su, and L. J. Guibas, ``PointNet++: Deep hierarchical feature learning on point sets in a metric space,'' \textit{arXiv Preprint}, Jun. 2017. [Online]. Available: \url{https://doi.org/10.48550/arXiv.1706.02413}

\bibitem{b12} 
C. R. Qi, H. Su, K. Mo, and L. J. Guibas, ``PointNet: Deep learning on point sets for 3D classification and segmentation,'' \textit{arXiv Preprint}, Apr. 2017. [Online]. Available: \url{https://doi.org/10.48550/arXiv.1612.00593}

\bibitem{b13}
Z. Xian, Y. Qiao, Z. Xu, T.-H. Wang, Z. Chen, \textit{et al.}, ``Genesis: A generative and universal physics engine for robotics and beyond,'' [Online]. Available: \url{https://genesis-embodied-ai.github.io/}

\bibitem{b14} 
J. Elsner, ``Taming the Panda with Python: A powerful duo for seamless robotics programming and integration,'' \textit{SoftwareX}, vol. 24, Art. no. 101532, 2023. [Online]. Available: \url{https://doi.org/10.1016/j.softx.2023.101532}

\end{thebibliography}
\end{document}